\documentclass{article}

\usepackage{arxiv}

\usepackage[utf8]{inputenc} % allow utf-8 input
\usepackage[T1]{fontenc}    % use 8-bit T1 fonts
\usepackage{hyperref}       % hyperlinks
\usepackage{url}            % simple URL typesetting
\usepackage{booktabs}       % professional-quality tables
\usepackage{amsfonts}       % blackboard math symbols
\usepackage{nicefrac}       % compact symbols for 1/2, etc.
\usepackage{microtype}      % microtypography
\usepackage{lipsum}
\usepackage{graphicx}
\usepackage{multirow}
\usepackage{authblk}

\usepackage[american]{babel}
\usepackage{natbib} % has a nice set of citation styles and commands
\usepackage{mathtools} % amsmath with fixes and additions
\usepackage{booktabs} % commands to create good-looking tables
\usepackage{subfigure}
\usepackage{hyperref}
\usepackage{makecell}

\usepackage{amsmath}
\usepackage{amssymb}
\usepackage{mathtools}
\usepackage{amsthm}
\newcommand\norm[1]{\Vert#1\Vert}

\newcommand\normF[1]{\Vert#1\Vert_F}

\newsavebox\affbox

\author[1]{Igor Meleshin}
\author[1,2]{Anna Chistyakova}
\author[2,3,4]{Anastasia Antsiferova}
\author[1,2,3]{Dmitriy Vatolin}
\affil[1]{Lomonosov Moscow State University, Moscow, Russia}
\affil[2]{ISP RAS Research Center for Trusted Artificial Intelligence, Moscow}
\affil[3]{MSU Institute for Artificial Intelligence, Moscow, Russia}
\affil[4]{Laboratory of Innovative Technologies for Processing Video Content, Innopolis University, Innopolis, Russia}
\affil[ ]{\texttt {\{igor.meleshin, aantsiferova, dmitriy\}@graphics.cs.msu.ru, a.chistyakova@ispras.ru}}
\date{}    

\title{Robustness as Architecture: Designing IQA Models to Withstand Adversarial Perturbations}
\begin{document}
\maketitle

\begin{abstract}
    Image Quality Assessment (IQA) models are increasingly relied upon to evaluate image quality in real-world systems --- from compression and enhancement to generation and streaming. Yet their adoption brings a fundamental risk: these models are inherently unstable. Adversarial manipulations can easily fool them, inflating scores and undermining trust. Traditionally, such vulnerabilities are addressed through data-driven defenses --- adversarial retraining, regularization, or input purification. But what if this is the wrong lens? What if robustness in perceptual models is not something to learn but something to design? In this work, we propose a provocative idea: robustness as an architectural prior. Rather than training models to resist perturbations, we reshape their internal structure to suppress sensitivity from the ground up. We achieve this by enforcing orthogonal information flow, constraining the network to norm-preserving operations --- and further stabilizing the system through pruning and fine-tuning. The result is a robust IQA architecture that withstands adversarial attacks without requiring adversarial training or significant changes to the original model. This approach suggests a shift in perspective: from optimizing robustness through data to engineering it through design.
\end{abstract}


\section{Introduction}\label{sec:intro}
Image quality assessment (IQA) plays a crucial role in image compression and processing, serving as a key tool for evaluating the perceptual quality of processed images. By providing objective quality estimates, IQA models reduce the need for costly and time-consuming subjective assessments. Depending on the availability of a reference image, IQA methods can be categorized into three approaches: full-reference (FR), reduced-reference (RR), and no-reference (NR). While FR-IQA compares a processed image with its original version, RR-IQA utilizes only partial information from the reference, and NR-IQA evaluates image quality without any reference, making it especially valuable for real-world applications such as image enhancement, denoising, and generation.

\begin{figure}[!t]
  \includegraphics[width=0.8\textwidth]{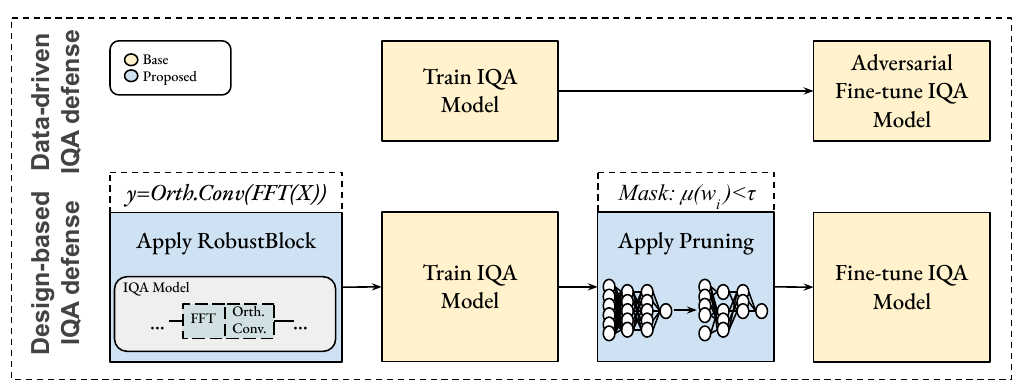}
  \centering
  \caption{Schematic pipelines comparing data-driven IQA defenses and the proposed design-based IQA defense. The data-driven approach relies on extensive training data to improve robustness, while the proposed design-based method leverages architectural modifications to enhance stability and resistance to adversarial attacks.}
  \label{fig:steps}
\end{figure}

As neural-network-based IQA models become  
increasingly adop\-ted in practical applications, ensuring their robustness against adversarial manipulations is essential~\citep{korhonen,zhang2022perceptualattacksnoreferenceimage,shumitskaya2022universalperturbationattackdifferentiable,shumitskaya2023fastadversarialcnnbasedperturbation}. Adversarial attacks on NR-IQA can exploit the lack of a reference image, making these models particularly vulnerable to manipulations. For instance, attackers can artificially boost image quality scores to achieve higher rankings in competitive benchmarks for image enhancement algorithms. This can lead to misleading performance evaluations and impact industry standards in image processing. Addressing these vulnerabilities is critical to maintaining the reliability of IQA in real-world scenarios.

The existing defense strategies for IQA models are \textit{data-driven} \citep{gushchin2024guardiansimagequalitybenchmarking}, as they primarily focus on modifying training procedures \citep{liu2024defense,advTrainingAnya} or perturbation purification \citep{advpurification}. 
In contrast to image classification, IQA is a regression task that requires robust models to remain highly sensitive to perceptual differences in images. Therefore, maintaining high performance in robust IQA models is more challenging than in classification. 
The drawback of adversarial training is that it is a significantly longer procedure involving the application of adversarial attacks to images. An adversarially trained model, consequently, is more robust only to attacks seen during training, which limits its robustness to unseen manipulations. Adversarial purification methods introduce significant computational overhead to model inference. For example, diffusion-based denoising offers a reasonable tradeoff between robustness and performance, albeit at the cost of being computationally expensive.
Overall, data-driven approaches, as they operate at the input or output level, leave IQA architectural modifications underexplored. 

Recent trends in related fields have shown that architecture choices effectively yield robustness in image classification~\citep{peng2023robarch}, but architectural defenses for IQA models are understudied. A more systematic investigation of architectural defenses could provide deeper insights into strengthening NR-IQA models against adversarial attacks.

Our study departs from \textit{data-focused} methods to explore \textit{design-based} robustness for NR-IQA models. Our main contributions are as follows:
\begin{itemize}
    \item We propose a novel architectural defense method based on orthogonalization, pruning, and fine-tuning techniques to enhance the adversarial robustness of NR-IQA models.
    \item We conduct a theoretical analysis of how the positioning of the orthogonalizing block affects the model's robustness.
    \item We perform experiments on two datasets and three NR-IQA models, evaluating the impact of our modifications on training speed, inference time, model performance, and robustness.
\end{itemize}

% Our code is publicly available at \textit{link hidden for blind review}.

\section{Related Work}

\subsection{Adversarial Attacks}
Adversarial examples are carefully designed inputs that introduce minor, imperceptible perturbations to images, causing a deep neural network to produce incorrect predictions with high confidence. Adversarial attacks on deep learning models have been extensively studied in image classification tasks \citep{goodfellow2014explaining,pgdattack, carlini2017evaluatingrobustnessneuralnetworks}.
Adversarial attacks targeting IQA models aim to manipulate the predicted quality scores of images without causing noticeable degradation in visual quality. One of the first attacks on IQA metrics is UAP \citep{shumitskaya2022universalperturbationattackdifferentiable}, which generates a universal adversarial noise that, when added to an input image, increases the IQA score regardless of the image's specific content. The perturbation is designed to be agnostic to the image to which it is applied, enabling it to affect a wide range of images similarly. A more advanced version of this attack, FACPA \citep{shumitskaya2023fastadversarialcnnbasedperturbation}, improves upon UAP by generating adversarial noise tailored to each image, leading to more effective attacks that both increase success rates and maintain higher perceptual quality. Ensuring that the added perturbations remain undetectable to human perception is one of the primary challenges in attacking IQA models. Korhonen et al. \citep{korhonen} proposed using a Sobel filter to conceal adversarial distortions in textured image areas. Zhang et al. \citep{zhang2022perceptualattacksnoreferenceimage} utilized various Full-Reference metrics, such as SSIM \citep{ssim}, LPIPS \citep{lpips}, and DISTS \citep{dists}, as additional terms in the objective function to control the imperceptibility of the added perturbations.

\subsection{Training-based Defenses}
Training-based defenses aim to enhance model robustness by modifying the training process through the use of specialized loss functions or incorporating adversarial examples. These methods improve resistance to adversarial attacks by explicitly regularizing model behavior or exposing it to perturbed inputs during training.

Gradient norm regularization (NT) \citep{liu2024defense} is a technique used to enhance the robustness of IQA models by penalizing large gradients in response to input perturbations. The idea is based on the observation that adversarial attacks often exploit large gradients to create perturbations that can significantly alter the model’s predictions. By regularizing the model's gradient, the training process forces the model to be less sensitive to small changes in input data, making it harder for adversarial attacks to succeed.

Adversarial training (AT) \citep{goodfellow2014explaining} is a popular approach for improving the robustness of deep networks by incorporating dynamically generated adversarial examples into the training set. However, base adversarial training is not directly applicable to IQA models. The challenge lies in adjusting the quality scores for attacked data, as there are no explicit labels for perturbed images. For IQA, adversarial training \citep{advTrainingAnya} involves modifying the subjective quality scores of the attacked images using full-reference metrics to prevent significant performance degradation on clean data.

\subsection{Architectural Adversarial Defenses}
One of the reasons for the vulnerability of neural networks to adversarial attacks is their architectural design. Therefore, many works propose modifying models to increase their robustness against adversarial perturbations. These techniques are primarily developed and evaluated for image classification tasks. In previous work \citep{xie2021smoothadversarialtraining}, the impact of the commonly used non-smooth activation function, ReLU, on classifier robustness is observed. The authors suggest replacing it with a more smooth alternative, such as ELU \citep{elu}, SiLU \citep{silu}, and GeLU \citep{gelu}. In another research \citep{Sqish}, a novel non-monotonic and smooth activation function called Sqish is introduced. It demonstrates improved robustness against a FGSM attack compared to ReLU.

Pruning was initially developed to reduce the computational cost of deep convolutional networks; however, it has also been explored as a defense mechanism against adversarial attacks \citep{jordao2021effect}. Researchers consider basic pruning methods, such as $l_1$-norm pruning and $l_2$-norm pruning, as well as more advanced techniques. For instance, pruning using Partial Least Squares (PLS) \citep{jordaopruning} projects the outputs of convolutional filters onto a latent space, where the importance of each filter is evaluated. Filters with low importance are iteratively removed, reducing redundancy while preserving critical features. An advanced variant builds upon this by utilizing a discriminative layer ranking criterion \citep{displspruning}. Instead of using a single subspace projection for all filters, this approach projects one subspace per layer, significantly reducing memory requirements. It focuses on reducing the depth of the network by removing entire layers rather than pruning filters across all layers, offering a more efficient way to optimize the network's structure. Other methods include pruning based on the Hilbert-Schmidt Independence Criterion (HSIC) \citep{hsicpruning}, which prunes a previously trained robust neural network while preserving adversarial robustness.

Additionally, modifications to network architecture inspired by biological mechanisms have been proposed \citep{dapello2020simulating}. For instance, introducing V1-like blocks mimicking the primary visual cortex has shown promise in enhancing model interpretability and robustness. A significant drawback of this method is the use of stochastic layers, which can negatively impact the model's convergence. The randomness introduced by these layers may lead to instability during training, making it more challenging to reach optimal performance. However, the authors demonstrated that their method provides a solid foundation for integration with other approaches, potentially mitigating convergence issues while enhancing the overall effectiveness of the model.

One of the methods that provides provable protection is the use of orthogonal convolutional blocks in the model \citep{trockman2021orthogonalizing}. The robustness of this method against adversarial attacks is based on the stability assessment for the Lipschitz function. Orthogonal convolution operators allow us to consider the entire model as a 1-Lipschitz function.
The construction of an orthogonal convolution is based on the Cayley transform \citep{trockman2021orthogonalizing}. The authors have demonstrated that their method is effective against adversarial attacks. However, experiments were conducted on small models since replacing all the original bundles with orthogonal ones significantly slows down the operation of the entire model. In addition, this protection method requires more memory for the model to function.

While the application of adversarial training for increasing IQA robustness has been previously studied (e.g., Chistyakova et al.~\citep{advTrainingAnya} proposed an enhanced adversarial training scheme), nobody has yet designed robust architectures in the IQA field. 

\section{Problem definition}
\textbf{Adversarial attack on IQA.} The goal of the adversarial attack against the IQA model is to increase the model's quality score of the original image by introducing small perturbations. Without loss of generality, attacks can be formulated to decrease the quality metric. Still, most of the literature focuses on attacks designed to increase the score, which is the approach we adopt. This can be formulated as:
\begin{equation}
    \max_{\|\delta\|_{p} \leq \varepsilon}\{f(x+\delta)-f(x)\}, \\
\end{equation}
where $f: X \rightarrow \mathbb{R}$ is the IQA model, $X\subseteq[0,255]^{3\times H \times W}$, $x\in X$ is the input image, $\delta$ is the perturbation, and $\varepsilon$ represents the radius of the $p$-norm ball, which defines the allowable perturbation magnitude, ensuring that the changes to the image remain imperceptible to the human eye.

\textbf{Architectural defense for IQA.} Architectural defense refers to strategies that improve the model's robustness to adversarial attacks by modifying its underlying architecture. 

Let $F(x) := F[f](x)$ represent the modified architecture of the original model $f(.)$, $\delta$ is adversarial perturbation bounded by $\varepsilon$ with a $p$-norm. The defense aims to ensure that the modified model is less sensitive to adversarial perturbations compared to the original model, which can be expressed as follows:
\begin{equation}
    \max_{\|\delta\|_{p} \leq \varepsilon}\{F(x+\delta)-F(x)\}<\max_{\|\delta\|_{p} \leq \varepsilon}\{f(x+\delta)-f(x)\}.\\
\end{equation}

At the same time, the modified model should maintain a correlation with subjective quality comparable to that of the original model.

\section{Proposed Method}

Limiting a model’s sensitivity to input perturbations is a key defense principle. From a theoretical perspective, it involves controlling the norm of activations to prevent adversarial amplification. To achieve this, weight matrices can be transformed using approximately orthogonal transformations, which preserve signal energy and stabilize forward and backward propagation. Orthogonalization is a natural choice here because it constrains the spectral norm of weight matrices, reducing sensitivity to input changes. However, it is not the only way—alternative approaches include spectral normalization, Jacobian regularization, or activation clipping. Orthogonalization, on the other hand, balances robustness with preserving signal detail particularly well in regression tasks, such as IQA.

In contrast to existing methods, we explore a \textit{design-based IQA architectural defense} strategy that directly limits the model’s sensitivity to input changes. Our pipeline (see Figure~\ref{fig:steps}) is structured as a modular procedure consisting of three core components:
\begin{enumerate}
    \item RobustBlock --- approximately orthogonal Fourier-domain transformations into the model to suppress adversarial amplification. We apply the FFT to shift the input into the frequency domain, where adversarial perturbations can be more easily isolated and suppressed via orthogonal operations. This also allows more efficient and stable implementation of convolutional transformations.
    \item Pruning --- a post-modification step that removes redundant or unstable channels, helping to restore efficiency and improve model stability.
    \item Fine-tuning --- a lightweight stage to recalibrate the model and recover perceptual performance after structural chan\-ges.
\end{enumerate}

This defense procedure is deliberately flexible: we do not restrict it to a specific block design but instead highlight a broader principle --- that robustness can be encouraged through the use of norm-preserving transformations. Together, these steps form a coherent, lightweight pipeline that enhances robustness against adversarial data while incurring minimal computational costs. While we demonstrate the effectiveness of each component --- orthogonalization, pruning, and fine-tuning --- we do not claim them to be optimal or exhaustive. Instead, we present this as a first step toward rethinking IQA robustness not as a post-hoc patch but as an architectural principle. We advocate for a shift toward designing defenses that are embedded in the model structure from the beginning --- integrated, efficient, and aligned with the model's perceptual function. Our work thus invites further exploration of how the robustness of IQA and other regression models can be achieved by design through a combination of geometric intuition, structural regularization, and minimal fine-tuning. 

\textbf{RobustBlock.} To balance between a model's performance and robustness, we suggest orthogonalizing certain layers in the network to suppress the impact of adversarial perturbations. This approach is inspired by certified defenses, which enforce Lipschitz continuity to guarantee bounded output changes under bounded input perturbations. However, existing methods typically require reparameterizing the entire network or computing tight bounds on the Lipschitz constant, often resulting in significant computational overhead. As a result, certified defenses remain largely impractical for high-dimensional perceptual tasks, such as IQA.

Motivated by this limitation, we propose a lightweight and flexible alternative: introducing approximate orthogonal transformations into selected layers of the network without modifying the overall architecture or requiring specialized training. These transformations --- such as orthogonal convolutions or structured norm-preserving mappings --- act as architectural priors that suppress perturbation amplification.

An analysis of the norm of the convolution transformation $\rho(x, y)=\|{x-y}\|$ illustrates the efficiency of orthogonal convolution in enhancing adversarial robustness. Convolution $conv(X)$ can be formulated as linear operator $\mathcal{W}: \mathbb{R}^{c_{in}n^2} \xrightarrow{} \mathbb{R}^{c_{out}n^2}$. Assuming $c_{in}=c_{out}=c$, the operator is represented by a matrix $W \in \mathbb{R}^{cn^2 \times cn^2}$. The input tensor $X \in \mathbb{R}^{n\times n \times c}$ is vectorized as $x \in \mathbb{R}^{n^2c}$. The perturbed input is $\tilde{x} \in \mathbb{R}^{n^2c}$, $ \tilde{x}=x+\delta$, where the perturbation $\delta \in \mathbb{R}^{n^2c}$ satisfies $\|\delta\|_2 \leq \varepsilon$, with $\varepsilon$ being the perturbation bound. 

If spectral norm satisfies $\|\mathcal{W}\|_2>1$, meaning that $\mathcal{W}$ expands at least one direction in the input space, then
$$\rho(\mathcal{W}\tilde{x}, \mathcal{W}x)=\|\mathcal{W}\delta\|>\|\delta\|,$$
\begin{equation}
\label{convolution_norm}
\|{\mathcal{W}\tilde{x}-\mathcal{W}x}\| >\|{\delta}\|,
\end{equation}
indicating increased sensitivity to perturbations. 

In contrast, when the convolution operator is norm-preserving --- as is the case for orthogonal transformations --- the Euclidean distance between perturbed and original inputs remains unchanged after the transformation. Formally, let $\mathcal{H}: \mathbb{R}^{c_{in}n^2} \rightarrow \mathbb{R}^{c_{out}n^2}$ be an orthogonal operator. Then:
\begin{equation}
\label{orthconvolution_norm}
|{\mathcal{H}\tilde{x}-\mathcal{H}x}| = |{\delta}|,
\end{equation}

which implies that the transformation does not amplify adversarial perturbations. This property makes orthogonal layers a natural fit for adversarially robust architectures. 

In practice, we can use an approach similar to that described in \citep{trockman2021orthogonalizing}. To make the convolution orthogonal, the input tensor must be square. This requirement arises because of the orthogonilizing methods. To satisfy this condition, we suggest using an adaptive pooling layer. As noted by \citep{trockman2021orthogonalizing}, for orthogonal convolution to be more efficient, the number of output channels should exceed the number of input channels. Based on this, we first apply a standard convolution to reduce the number of channels before applying the orthogonal one.

\begin{table*}[t]
\caption{SROCC$\uparrow$, PLCC$\uparrow$, AbsGain$\downarrow$, and R-Score$\uparrow$ values for the Linearity IQA model with different positions of the RobustBlock. In the Modification column, “RobustBlock N” indicates that the RobustBlock is inserted before the N-th convolutional block in the model. The InputDim column specifies the tensor dimensions fed into the RobustBlock. Training epoch time and inference time on the test set were measured using an NVIDIA Tesla A100 80 GB GPU.}
\label{tab:cayley-performance}
\vskip 0.15in
\begin{center}
\begin{small}
\begin{sc}
\begin{tabular}{ccccccccccc}
\toprule
Modification & InputDim & \multicolumn{2}{c}{Performance} & \multicolumn{2}{c}{Robustness} & \multicolumn{2}{c}{Time}  \\
 & & SROCC$\uparrow$ & PLCC$\uparrow$ & AbsGain$\downarrow$ & RScore$\uparrow$ & train. epoch & test  \\
\midrule
 --- & --- & \textbf{0.926} & \textbf{0.936} & 0.409	& 0.012 & \textbf{100.727} & \textbf{31.735} \\
 AT & --- & 0.855 & 0.877 & \textbf{-0.209} & \textbf{0.025} & 183.682 & \textbf{32.216} \\
 NT & --- & \textbf{0.904} & \textbf{0.921} & \textbf{0.265} & \textbf{0.026} & 229.612 & \textbf{33.028} \\
 RobustBlock 1 & $3\times498\times664$ & 0.766 & 0.794 & 0.415 & 0.018 & \textbf{131.172} & 36.787 \\
 RobustBlock 2 & $64\times125\times166$ & 0.793 & 0.823 & 0.508 & 0.006 & \textbf{129.313} & 63.864 \\
 RobustBlock 4 & $128\times63\times83$ & 0.854 & 0.876 & 0.402 & 0.016 & 199.581 & 103.655 \\
 RobustBlock 6 & $1024\times32\times42$ & \textbf{0.923} & \textbf{0.936} & \textbf{0.289} & \textbf{0.031} & 181.826 & 89.703 \\
\bottomrule
\end{tabular}
\end{sc}
\end{small}
\end{center}
\vskip -0.1in
\end{table*}

\begin{figure}[!htb]
    \centering
    \includegraphics[width=\linewidth]{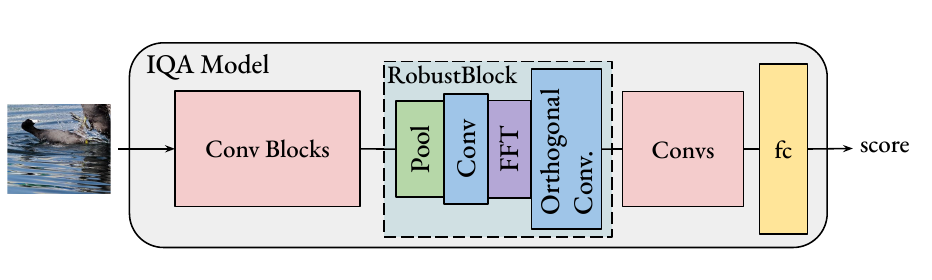} 
    \caption{Proposed modification of the model's architecture for adversarial attack defense.}
    \label{fig:arch}
\end{figure}

As a result of the above reasoning, we propose the \textit{RobustBlock}, which is shown in Figure~\ref{fig:arch}. The RobustBlock includes a \textit{pooling layer} that reshapes the tensor into a square form, a convolutional layer that reduces the number of channels, and a \textit{Fast Fourier Transform (FFT)} followed by an orthogonal convolution. Performing the orthogonal convolution in the Fourier domain leverages the accumulation of noise in the frequency space and allows efficient operations on circular matrices as well as more efficient computation of inverse convolutions.

\textbf{RobustBlock Position.} To determine the location for the orthogonal convolution in the original model, we derived the following conclusions.
\label{sec:block_position}

Consider an adversarial perturbation $\delta \in \mathbb{R}^{n\times 1}: \|\delta\|_2\leq\varepsilon$ for the linear operator $\mathcal{W}: \mathbb{R}^{n} \xrightarrow{} \mathbb{R}^{m}$  
 with input $x \in \mathbb{R}^{n\times 1}$ and $\Tilde x = x + \delta$. Then, we have an adversarial attack as the following optimization problem:
\begin{equation}
\label{eq:conv_attack}
\max_{\|\delta\|_2\leq\varepsilon} \|\mathcal{W}\Tilde x - \mathcal{W} x \|. \\
\end{equation}
This is equivalent to the condition that the perturbation vector is a singular vector.

\textbf{Lemma 1.} \textit{If linear operator $\mathcal{W}: \mathbb{R}^{n} \xrightarrow{} \mathbb{R}^{m}$, $\frac{\normF{\mathcal{W}}}{\norm{\mathcal{W}}_2}>\frac{m}{n}$, $m\ge n$, $\norm{\mathcal{W}}_2>1$ and orthogonal linear operator $\mathcal{H}: \mathbb{R}^{n} \xrightarrow{} \mathbb{R}^{n}$, perturbation vector $\delta \in \mathbb{R}^{n}$, $\{\norm{\delta}\leq\varepsilon\}$ for $x \in \mathbb{R}^n$. $\delta=\alpha v_1$, where $v_1$ is right singular vector of linear operator $\mathcal{W}$ corresponding to the maximum singular value.
Then $\norm{\mathcal{W}\mathcal{H}(\tilde{x}-x)}>\frac{m}{n}\norm{\delta}$.}

The proof of the lemma is provided in the supplementary material.

Now, using Eq.~\ref{eq:conv_attack} and Lemma 1, we represent the input image with $c_{in}$ channels and $s_{in}\times s_{in}$ dimension as vector $X \in \mathbb{R}^{c_{in}s_{in}^2\times 1}$. Let the convolution and semi-orthogonal convolution be represented as $H(X) \in \mathbb{R}^{c_{out0}s_{out}^2\times 1}$, $W(X) \in \mathbb{R}^{c_{out}s_{out}^2\times 1}$, respectively. We assume the following: $\norm{H(X)}=\norm{X}$, $\frac{\normF{W}}{\norm{W}_2}>\frac{c_{out}s_{out}^2}{c_{in}s_{in}^2}$, $c_{out}s_{out}^2\ge c_{in}s_{in}^2$. For fixed convolution parameters (kernel size, padding, stride, and dilation), we can represent the sizes of the output tensors as suggested in \citep{dumoulin2016guide}:
$$s_{out}=\lfloor \frac{s_{in}+2*pad.-(ker.-1)*dilation-1}{stride}+1 \rfloor$$
This can be rewritten as: 
\begin{equation}
    \begin{aligned}
    & s_{out}=\lfloor \eta s_{in}-\mu \rfloor, \text{ where } \eta=\frac{1}{stride},\\ 
    & \mu=\frac{(ker.-1)*dilation+1-2*pad.}{stride}-1.
    \end{aligned}
\end{equation}

If we use semi-orthogonal convolution and ordinary convolution as a composition of functions, we can estimate the perturbation as follows:
\begin{equation}
\label{eq:channels}
\begin{aligned}
\|W(H(X+\varepsilon))-W(H(X))\|&>\frac{c_{out}s_{out}^2}{c_{in}s_{in}^2}\|\varepsilon\| \\
&=\frac{c_{out}}{c_{in}}\Bigl( \eta - \frac{\mu}{s_{in}} \Bigr)^2.
\end{aligned}
\end{equation}
From this inequality, we can conclude that the smaller the input image dimension, the higher the disturbance threshold obtained after convolution.

Based on the above reasoning, it is better to place the RobustBlock closer to the fully connected layer of the model because, at that stage, the ratio of the output tensor dimension to the input dimension is smaller. We test the various positions of the RobustBlock within the model to investigate its effect on learning time and inference and discuss the results in Section~\ref{sec:results}.

\textbf{Pruning and Fine-Tuning.} To make the model lighter and more stable, we propose applying pruning. Pruning allows us to remove less important weights based on a set of criteria $\mu$. Masking process can be formulated as:
$\mu(w_i) < \tau,$ where $w_i$ is a weight column and $\tau$ is the threshold value for masking. After pruning the weights, we add a fine-tuning step to maintain a high level of performance in the final model. The entire process is illustrated in Figure ~\ref{fig:steps}.

\section{Experiments}
\label{sec:experiments}
\textbf{Models.} We evaluated our defense strategies on three IQA models: Linearity \citep{li2020norm}, TReS \citep{golestaneh2022no}, DBCNN \citep{dbcnn}, and KonCept \citep{hosu2020koniq}. KonCept \citep{hosu2020koniq}, designed for the KonIQ-10k dataset, is built on the Inception-ResNet-V2 backbone, pre-trained on ImageNet, and is used for image quality assessment. Linearity \citep{li2020norm}  is based on the ResNet architecture. Input data is processed through the last two convolutional blocks, creating two separate branches. Each branch is passed through encoder blocks, after which the outputs are concatenated and fed into a fully connected layer. The key feature of this model is the use of a norm-in-norm error, which ensures faster convergence during training. DBCNN \citep{dbcnn} employs a branching architecture with two blocks: VGG16 and S-CNN. The S-CNN block is specifically trained on data with various noise levels. After processing through these blocks, the output features are combined using bilinear pooling and passed to a fully connected layer. TReS \citep{golestaneh2022no} is a transformer-based built on the ResNet50 backbone, pre-trained on ImageNet. The outputs from the ResNet convolutional layers are used as input features for the transformer encoder.

% For comparison with the transformer models, we chose the TReS \citep{golestaneh2022no}. 
% Since transformer-based metrics are more robust to attacks due to their constructive design, we also compared the proposed method with the TReS model \citep{golestaneh2022no}.

\textbf{Datasets.} To train our models, we used the KonIQ-10k dataset \citep{hosu2020koniq}, which contains 10,074 images with a resolution of $1024 \times 768$, paired with subjective quality values based on Mean Opinion Scores (MOSs). Following the original Linearity training procedure, we resized the images to $664 \times 498$ for all models to ensure a fair comparison. The dataset was divided into 70\% for training, 10\% for validation, and 20\% for testing. Additionally, to evaluate robustness, we used the NIPS2017 dataset \citep{nips-dataset}, which contains 1,000 images with a resolution of $299 \times 299$. No additional transformations were applied to the images. This dataset is commonly used for attack development and testing and omits MOSs.

\textbf{Performance evaluation.} To measure the correlation of IQA model scores with subjective ratings, we used the Spearman rank order correlation coefficient (SROCC) and Pearson linear correlation coefficient (PLCC).

\textbf{Robustness evaluation.} To evaluate the robustness of IQA models, we used Projected Gradient Descent (PGD) attack \citep{pgdattack}, Universal Adversarial Perturbations (UAP) \citep{moosavi2017universal}, Spatially transformed adversarial examples (stAdv) \citep{xiao2018spatially}. These methods are widely used in defense development studies as it allows for simple control over the strength of the attack by adjusting the perturbation magnitude, and it offers precise optimization by varying the number of iterations. Robustness was quantified using Absolute Gain and Robustness score (R-Score) \citep{zhang2022perceptualattacksnoreferenceimage}. Given a model $f$, clear data $x_i$ and perturbed data $\tilde{x_i}$, these metrics are defined as follows:
\begin{equation}
    Abs.Gain=\frac{1}{n}\sum^{n}_{i=1}\left(f(\tilde{x_i})-f(x_i)\right),\\
\end{equation}
\begin{equation}
    R_{score}=\frac{1}{n}\sum^{n}_{i=1}\log\left(\frac{\max\{\beta_1-f(\tilde{x_i}), f(x_i)-\beta_0\}}{|f(\tilde{x_i})-f(x_i)|}\right), \\
\end{equation}

\begin{table*}[t]
\caption{SROCC$\uparrow$, PLCC$\uparrow$, AbsGain$\downarrow$, and R-Score$\uparrow$ of IQA models with different architectures. The original models are compared with those incorporating a RobustBlock, denoted as +orth. AbsGain and R-Score values are calculated for images with one-step PGD attack. The best value for each model is highlighted in bold.}
\label{tab:all-cayley-perfomance}
\vskip 0.15in
\begin{center}
\begin{small}
\begin{sc}
\begin{tabular}{lcccccccccc}
\toprule
Model & \multicolumn{2}{c}{Performance} & \multicolumn{2}{c}{Robustness} \\
 &  SROCC$\uparrow$ & PLCC$\uparrow$ & AbsGain$\downarrow$ & RScore$\uparrow$ \\
\midrule
KonCept & 0.915 & 0.929 & 0.359 & 0.013 \\
KonCept+orth. & \textbf{0.920} & \textbf{0.932}& \textbf{0.301} & \textbf{0.015} \\
\midrule
Linearity & \textbf{0.926} & \textbf{0.936} & 0.409 & 0.012 \\
Linearity+orth. & 0.923 & \textbf{0.936} & \textbf{0.289} & \textbf{0.031} \\
\midrule
DBCNN & \textbf{0.873} & \textbf{0.885} & 0.260 & 0.033 \\
DBCNN+orth. & 0.838 & 0.852 &\textbf{ 0.218} & \textbf{0.068} \\
\midrule
TReS & 0.918 & \textbf{0.929} & \textbf{0.300} & 0.027 \\
TReS+orth. & \textbf{0.919} & \textbf{0.929} & 0.314 & \textbf{0.029} \\
\bottomrule
\end{tabular}
\end{sc}
\end{small}
\end{center}
\vskip -0.1in
\end{table*}

\begin{table*}[ht]
\caption{Performance and robustness metrics of the Linearity model and our method with different orthogonal convolution implementations. The best value for each metric within each method group is highlighted in bold.}
\label{tab:linearity-orthogonal-convolution}
\vskip 0.15in
\begin{center}
\begin{small}
\begin{sc}
\begin{tabular}{lrrrrrrrrrr}
\toprule
\multicolumn{1}{c}{\multirow{2}{*}{\textbf{model}}} & \multicolumn{1}{c}{\multirow{2}{*}{\textbf{SROCC$\uparrow$}}} & \multicolumn{1}{c}{\multirow{2}{*}{\textbf{PLCC$\uparrow$}}} & \multicolumn{2}{c}{\textbf{AbsGainAUC$\downarrow$}}                                 & \multicolumn{2}{c}{\textbf{RScoreAUC$\uparrow$}}                                  & \multicolumn{2}{c}{\textbf{AbsGain$\downarrow$}}                                  & \multicolumn{2}{c}{\textbf{RScore$\uparrow$}}                                   \\
\multicolumn{1}{c}{}                                & \multicolumn{1}{c}{}                                & \multicolumn{1}{c}{}                               & \multicolumn{1}{c}{\textbf{PGD-1}} & \multicolumn{1}{c}{\textbf{PGD-8}} & \multicolumn{1}{c}{\textbf{PGD-1}} & \multicolumn{1}{c}{\textbf{PGD-8}} & \multicolumn{1}{c}{\textbf{UAP}} & \multicolumn{1}{c}{\textbf{stAdv}} & \multicolumn{1}{c}{\textbf{UAP}} & \multicolumn{1}{c}{\textbf{stAdv}} \\
\midrule
Linearity                                           & \textbf{0.926}                                      & \textbf{0.936}                                     & 0.409                              & 1.821                              & 0.013                              & 0.001                              & 0.465                            & 0.025                              & 0.463                            & 1.732                              \\
+Cayley                                             & 0.921                                               & 0.935                                              & \textbf{0.261}                     & \textbf{1.481}                     & \textbf{0.034}                     & 0.001                              & 0.360                            & \textbf{0.022}                     & 0.668                            & \textbf{1.862}                     \\
+AOC                                                & \textbf{0.923}                                      & \textbf{0.937}                                     & 0.312                              & 1.540                              & 0.026                              & 0.001                              & \textbf{0.359}                   & \textbf{0.023}                     & \textbf{0.716}                   & 1.780                              \\
+AOL                                                & \textbf{0.926}                                      & \textbf{0.936}                                     & \textbf{0.305}                     & \textbf{1.499}                     & \textbf{0.027}                     & 0.001                              & \textbf{0.358}                   & \textbf{0.023}                     & \textbf{0.717}                   & \textbf{1.789}  \\
\bottomrule
\end{tabular}
\end{sc}
\end{small}
\end{center}
\vskip -0.1in
\end{table*}

where $\beta_0$ and $\beta_1$ are the minimum and maximum values of metric output, respectively.
Absolute Gain measures the average absolute change in the predicted quality after the attack. R-Score quantifies the robustness of the model by evaluating the ratio between the maximum allowable change in quality prediction and the actual change, averaged over all attacked images and expressed in a logarithmic scale.

 For our experiments with PGD attack, we chose perturbation levels $\varepsilon = \left\{2, 4, 6, 8, 10\right\}/255$, iteration counts $\left\{1, 8 \right\}$, and a step size of $1/255$. Since we used several perturbation levels for the PGD attack, we calculated the area under the AbsGain curve (AdsGainAUC) and the R-Score (R-ScoreAUC) to assess the robustness against this perturbation. UAP and stAdv generated 10 and 5 iterations, respectively.

For each predicted score, we applied min-max normalization using the IQA model’s score range from the training set. The final results are presented as the integral of metric-$\varepsilon$ curves, averaged over all test dataset images. Since the selected KonIQ-10k and NIPS datasets have a 2:1 ratio, we average the model defense scores over these two datasets with the specified coefficients in our tests.

\subsection{Implementation Details}
Our experiments were performed using an Nvidia Tesla A100 80-Gb GPU.

When implementing pruning, we encountered some characteristics of various methods. Some pruning techniques, such as PLS and Distillation PLS, require a certain amount of training data to generate the weight matrix for the projection transformation. We selected 200 samples, which is approximately 3\% of the total dataset. For all pruning methods, we used pre-trained models and applied pruning and fine-tuning over 5 epochs. In our implementation of the RobustBlock, we rely on several key elements:

\textbf{Pooling layer.} We use the adaptive pooling method, which transforms an input tensor with dimension $[ C, H, W ]$, where $H \neq W$, into a tensor with size $$[C, min(H,W),  min(H,W)].$$ This allows us to compress data efficiently without significant loss of information.

\textbf{Convolutional layer.} This layer is essential for reducing the number of channels before performing orthogonal convolution. We set the number of output channels in the intermediate convolution to be half the number of input channels. This allowed us to significantly reduce memory usage, which was especially important in some cases, as orthogonal convolution requires relatively large memory resources. According to \citep{trockman2021orthogonalizing}, it can be argued that increasing the number of channels contributes to the maximum efficiency of orthogonal convolution.

As a result, the RobustBlock can be compared to the Bottleneck layer, as it reduces the number of channels during operation. However, at the output, the block restores the original number of channels, and the output tensor becomes square.

For IQA models Linearity and KonCept, we placed the RobustBlock before the last block of convolutions. Since the DBCNN metric splits into two branches during its operation, we added a RobustBlock to both branches before the previous block of convolutions.

\section{Results}
\label{sec:results}
\begin{figure}[!tb]
\centering
\includegraphics[width=0.5\linewidth]{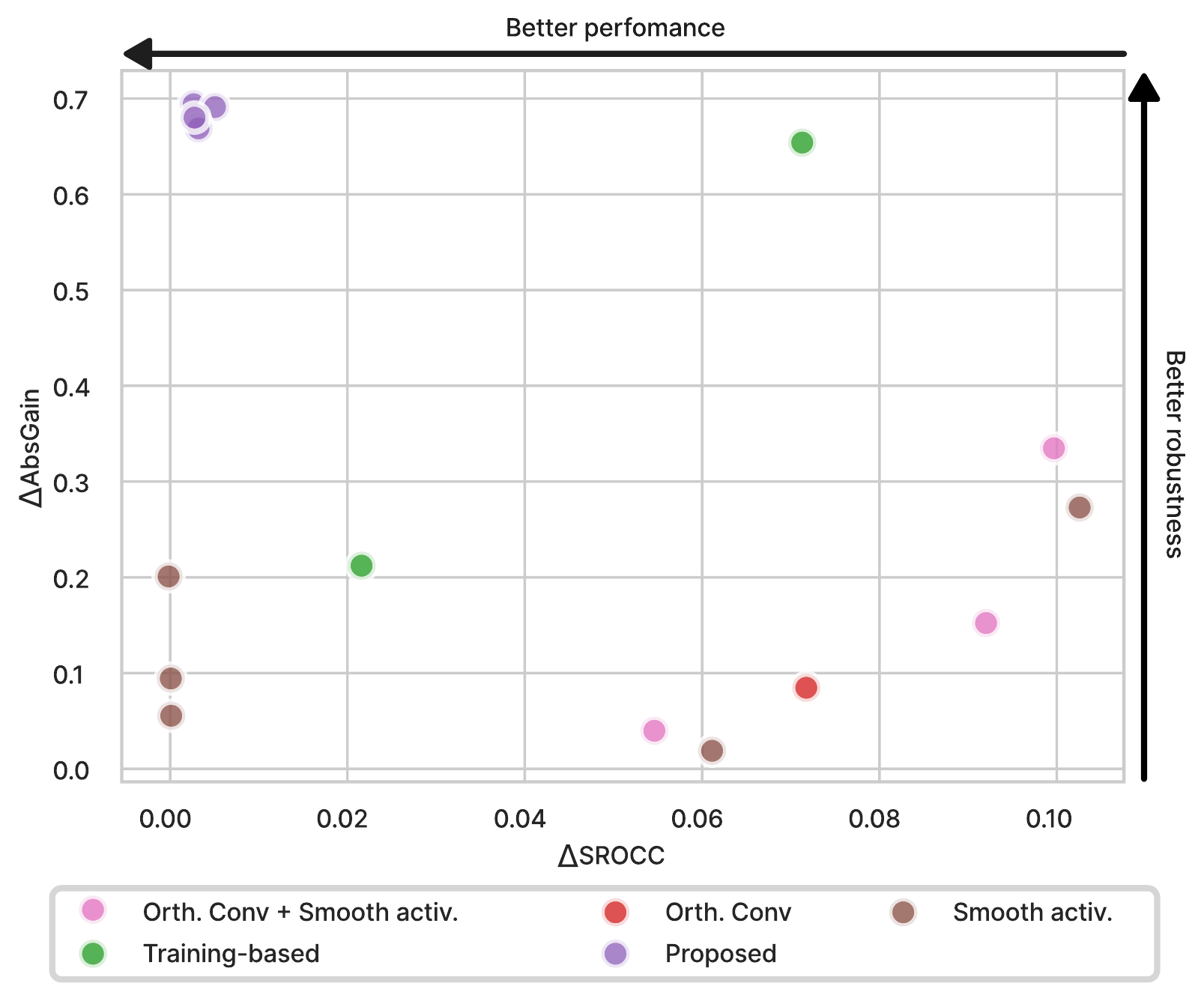}
% \caption{Deltas of AbsGainAUC and SROCC relative to the original IQA model for various defenses on the NIPS2017 dataset. The proposed method outperforms in terms of robustness while maintaining a high SROCC.}
\caption{Trade-off between robustness and quality correlation on the NIPS2017 dataset. Our architectural modification (purple) consistently reduces vulnerability to adversarial attacks (lower $\mathrm{AbsGainAUC}$) while maintaining high perceptual alignment (minimal drop in $\mathrm{SROCC}$). This highlights that architectural priors can realize robustness–performance trade-offs that training-based methods often fail to achieve.}
\label{fig:initial-graph}
\end{figure}
\subsection{Ablation Study}
\textbf{Block position.} To support our theoretical analysis in Section~\ref{sec:block_position}, we conducted an experiment using various positions of the RobustBlock in the model. The results, presented in Table~\ref{tab:cayley-performance}, confirm that when the ratio of output channels $c_{out}$ to input channels $c_{in}$ in the convolutional layer preceding the RobustBlock is lower, the model becomes more robust to adversarial attacks. Additionally, when the modification is applied closer to the beginning of the model, the pre-trained weights of the feature encoder are more disrupted, leading to a significant decrease in correlation with subjective quality scores. Based on both theoretical analysis and empirical evidence, we found that the optimal position for the RobustBlock is between the last two blocks of the convolutions, near the fully connected layer. For all subsequent experiments, we used this placement unless otherwise stated.

\textbf{Orthogonalizing transformation.} To assess the effectiveness of orthogonal convolutions, we tested this approach on three different IQA models. The results, presented in Table~\ref{tab:all-cayley-perfomance}, show that RobustBlock improves model robustness against adversarial attacks and, in some cases, can even enhance overall performance.
To demonstrate that our protection method is effective for different approaches to orthogonalization, we conducted tests using different types of orthogonal convolutions. Table~\ref{tab:linearity-orthogonal-convolution} shows the results of Linearity when using Adaptive Orthogonal Convolution (AOC) \citep{boissin2025adaptive}, Almost-Orthogonal Layer (AOL) \citep{prach2022almost} convolutions and Cayley convolution \citep{trockman2021orthogonalizing}.
As a result, we chose the Cayley transformation as the main method of orthogonalization and in further experiments we will use this implementation.

\textbf{Pruning.} RobustBlock introduces additional computational overhead, as detailed in Table~\ref{tab:cayley-performance}. To mitigate this, we incorporate pruning. Originally, pruning methods were developed to speed up model inference; however, they have also been found to enhance model robustness against adversarial attacks \citep{jordao2021effect}. As part of our ablation study, we conducted experiments using pruning as the sole defense mechanism. Detailed results of these experiments are provided in the supplementary material. However, the results did not show any significant improvements in robustness. An essential factor in pruning is selecting the optimal percentage of weights to remove. We determined that pruning 10\% of weights is a generally suitable choice for all evaluated IQA models. Despite this relatively low pruning rate, we observed a decline in correlation with subjective quality scores, prompting us to apply additional fine-tuning. However, even after further training, some IQA models still exhibited a significant decrease in correlation with
subjective quality scores.  

% Pruning is a method aimed at increasing the speed and accuracy of the model. As part of our research, we have tested several pruning approaches, and the results of these tests are presented in Table~\ref{tab:performance-prune}.
% It follows from these data that pruning is an effective tool for protecting models, especially in combination with other methods. However, by itself, it can only withstand minor attacks.
% Another important aspect of using pruning is choosing the optimal percentage of weights to remove. We settled on the value of 10\%, as it turned out to be universal for all metrics. Despite our choice of a small percentage of weights, the correlation was still decreasing, so we turned to further training. However, even with additional training, the decrease in correlation was too significant for some of the tested models.

\begin{table*}[ht]
\caption{Performance and robustness metrics of IQA models with different architectural variations. AbsGain and R-Score are shown for UAP and stAdv, while area-under-curve (AUC) values are provided for PGD attacks with 1 and 8 iterations. The best result for each metric within each model group is highlighted in bold.}
\label{tab:iterative-cayley-robustness}
\vskip 0.15in
\begin{center}
\begin{small}
\begin{sc}
\begin{tabular}{lrrrrrrrrrr}
\toprule
\multicolumn{1}{c}{\multirow{2}{*}{\textbf{model}}} & \multicolumn{1}{c}{\multirow{2}{*}{\textbf{SROCC$\uparrow$}}} & \multicolumn{1}{c}{\multirow{2}{*}{\textbf{PLCC$\uparrow$}}} & \multicolumn{2}{c}{\textbf{AbsGainAUC$\downarrow$}}                                 & \multicolumn{2}{c}{\textbf{RScoreAUC$\uparrow$}}                                  & \multicolumn{2}{c}{\textbf{AbsGain$\downarrow$}}                                  & \multicolumn{2}{c}{\textbf{RScore$\uparrow$}}                                   \\
\multicolumn{1}{c}{}                                & \multicolumn{1}{c}{}                                & \multicolumn{1}{c}{}                               & \multicolumn{1}{c}{\textbf{PGD-1}} & \multicolumn{1}{c}{\textbf{PGD-8}} & \multicolumn{1}{c}{\textbf{PGD-1}} & \multicolumn{1}{c}{\textbf{PGD-8}} & \multicolumn{1}{c}{\textbf{UAP}} & \multicolumn{1}{c}{\textbf{stAdv}} & \multicolumn{1}{c}{\textbf{UAP}} & \multicolumn{1}{c}{\textbf{stAdv}} \\
\midrule
Linearity                                           & \textbf{0.926}                                      & \textbf{0.936}                                     & 0.409                              & 1.821                              & 0.013                              & 0.001                              & 0.465                            & 0.025                              & 0.463                            & 1.732                              \\
+NT                                        & 0.904                                               & 0.921                                              & \textbf{0.305}                     & \textbf{1.479}                     & \textbf{0.026}                     & 0.002                              & \textbf{0.190}                   & \textbf{0.015}                     & \textbf{0.825}                   & \textbf{1.937}                     \\
+AT                                        & 0.855                                               & 0.877                                              & 0.365                              & \textbf{1.373}                     & \textbf{0.026}                     & 0.002                              & \textbf{0.322}                   & \textbf{0.013}                     & 0.491                            & \textbf{1.985}                     \\
+our                                       & \textbf{0.921}                                      & \textbf{0.935}                                     & \textbf{0.261}                     & 1.481                              & \textbf{0.034}                     & 0.001                              & 0.360                            & 0.022                              & \textbf{0.668}                   & 1.862                              \\
\midrule
KonCept                                             & \textbf{0.915}                                      & \textbf{0.929}                                     & 0.359                              & 1.190                              & 0.015                              & 0.002                              & 1.276                            & \textbf{0.006}                     & -0.293                           & 2.176                              \\
+NT                                          & 0.794                                               & 0.808                                              & \textbf{0.122}                     & \textbf{0.250}                     & 0.036                              & \textbf{0.024}                     & \textbf{0.419}                   & \textbf{0.001}                     & \textbf{0.608}                   & \textbf{3.873}                     \\
+AT                                          & 0.815                                               & 0.850                                              & \textbf{0.062}                     & \textbf{0.078}                     & \textbf{0.046}                     & \textbf{0.043}                     & 0.633                            & 0.009                              & \textbf{0.488}                   & \textbf{2.386}                     \\
+our                                         & \textbf{0.884}                                      & \textbf{0.908}                                     & 0.244                              & 0.534                              & \textbf{0.040}                     & 0.011                              & \textbf{0.396}                   & \textbf{0.006}                     & 0.413                            & 2.324                              \\
\midrule
TReS                                                & \textbf{0.918}                                      & \textbf{0.929}                                     & \textbf{0.300}                     & 1.086                              & \textbf{0.027}                     & 0.002                              & 0.472                            & 0.103                              & 0.181                            & 0.862                              \\
+NT                                             & 0.887                                               & 0.901                                              & \textbf{0.307}                     & \textbf{0.748}                     & \textbf{0.029}                     & \textbf{0.015}                     & \textbf{0.273}                   & \textbf{0.122}                     & \textbf{0.395}                   & \textbf{1.427}                     \\
+AT                                             & 0.886                                               & 0.904                                              & 0.357                              & 1.085                              & 0.026                              & \textbf{0.008}                     & \textbf{0.426}                   & \textbf{0.117}                     & \textbf{0.188}                   & 0.846                              \\
+our                                            & \textbf{0.919}                                      & \textbf{0.929}                                     & 0.314                              & \textbf{1.084}                     & \textbf{0.029}                     & \textbf{0.008}                     & 0.503                            & 0.051                              & 0.178                            & \textbf{0.903} \\
\bottomrule
\end{tabular}
\end{sc}
\end{small}
\end{center}
\vskip -0.1in
\end{table*}

\subsection{Overall results}
Our final experiments employed the most effective strategies identified in our ablation study. Our method consists of RobustBlock placed near the fully connected layers, followed by pruning 10\% of the weights (using $l_2$-pruning for Linearity and PLS-pruning for KonCept). Finally, the model undergoes fine-tuning for five epochs to restore high correlation with subjective quality scores.

\textbf{Impact on performance.} In Figure~\ref{fig:initial-graph}, we compare our proposed defense method with existing approaches, including specially designed training-based methods for IQA models. The performance decrease for our method is less than 1\% compared to the original value. Our approach outperforms other methods, except for those that employ smoothed activation functions, which do not provide strong adversarial robustness.

% We have considered various variations of protection methods, including various implementations of the proposed method. The results are illustrated in Figure~\ref{fig:initial-graph}. The numerical results are presented in Table~\ref{tab:initial-figure}.

\textbf{Impact on robustness.} Figure~\ref{fig:initial-graph} shows that our proposed me\-thod outperforms all other techniques in terms of robustness against the one-step PGD attack. In Table~\ref{tab:iterative-cayley-robustness}, we apply our proposed method to three convolutional IQA models. The results demonstrate the effectiveness of the method for all tested models in terms of RScore and AbsGain for PGD, UAP, stAdv attacks with varying magnitudes and numbers of iterations.

% Additionally, Table~\ref{tab:iterative-cayley-robustness} includes results for the transformer-based model TReS \citep{golestaneh2022no}, which is inherently more robust than convolutional models due to its design. 
% The results indicate that our method enhances the robustness of convolutional models beyond that of TReS in nearly all cases.
Our experimental results, as presented in Table~\ref{tab:iterative-cayley-robustness}, demonstrate that the proposed architectural defense improves the adversarial robustness of IQA models. Notably, our method consistently achieves robustness scores under all selected attacks compared to baseline approaches. Our method maintains a strong correlation with subjective human scores, as evidenced by SROCC and PLCC across all tested datasets. These findings highlight that our design-based modifications not only suppress adversarial sensitivity but also preserve the model’s ability to accurately reflect human visual judgments, making our approach highly effective for real-world IQA applications where both robustness and perceptual alignment are critical.

\section{Discussion}
% \begin{table}
% \caption{SROCC$\uparrow$, AbsGain$\downarrow$ of the Linearity IQA model trained on the KonIQ-10k dataset using training-based methods with different activation functions. AbsGain is calculated for a one-step PGD attack. The best value is highlighted in bold for each training-based method separately.
% }
% \label{tab:performance-train-activation}
% \vskip 0.15in
% \begin{center}
% \begin{small}
% \begin{sc}
% \begin{tabular}{lcccc}
% \toprule
% Method & Activation & SROCC$\uparrow$ & AbsGain$\downarrow$\\
% \midrule
% original & relu & 0.926 & 0.480 \\
% \midrule
% NT & relu & \textbf{0.904} & 0.311 \\
%  & Felu & 0.863 & \textbf{0.278} \\
%  & Fgelu & 0.819 & \textbf{0.067} \\
%  & Fsilu & 0.811 & 0.446 \\
%  & elu & \textbf{0.915} & \textbf{0.227} \\
%  & gelu & 0.733 & 0.454 \\
%  & silu & \textbf{0.917} & 0.286 \\
% \midrule
% AT & relu & \textbf{0.855} & \textbf{-0.197} \\
%  & Felu & 0.845 & \textbf{-0.171} \\
%  & Fgelu & 0.767 & -0.067 \\
%  & Fsilu & 0.753 & 0.407 \\
%  & elu & 0.852 & -0.112 \\
%  & gelu & \textbf{0.871} & -0.039 \\
%  & silu & \textbf{0.858} & \textbf{-0.169} \\
% \bottomrule
% \end{tabular}
% \end{sc}
% \end{small}
% \end{center}
% \vskip -0.1in
% \end{table}
In Section~\ref{sec:results}, we noted that smoothed activation functions help preserve correlation with subjective quality assessments but do not provide strong robustness against adversarial attacks. Given this, we decided to investigate their impact separately. We conducted a series of experiments to evaluate various smoothed activation functions. We used two strategies for activation replacement: complete replacement and partial replacement. Full replacement involves substituting all ReLU activations in the original model with an alternative activation function, whereas partial replacement targets only those activations in layers that have not been pre-trained.
Our analysis revealed that complete replacement of the activation function consistently enhances robustness --- in some cases, it even reduces correlation with subjective quality assessments. In contrast, partial replacement can improve model stability without compromising performance. This approach is ineffective for IQA models, where only a small number of layers are added to the feature encoder. Replacing the activation function is not always an effective solution. A key finding from our work is that smoother activation functions contribute to increased robustness, and these modifications should be applied to layers that have not undergone pre-training.

However, combining training-based defense methods with smoo\-thed activation functions has demonstrated effectiveness in some cases. Thus, while the complete replacement of an activation function is not highly effective as a standalone architectural defense approach, it can serve as a valuable tool to enhance training-based defenses.

The detailed results of these additional experiments are provided in the supplementary material.

\section{Conclusion} 
In this work, we proposed a rethinking of adversarial robustness in IQA: instead of learning robustness from data, we design it into the architecture. Our defense strategy replaces the typical data-centric pipeline with a modular architectural prior — introducing orthogonal transformations (RobustBlock), stability-enhancing pruning, and lightweight fine-tuning. We theoretically and empirically demonstrated that such design-based robustness can suppress adversarial sensitivity while maintaining high correlation with human perception. This perspective opens up a new design paradigm for multimedia systems: robustness not as a by-product of training, but as a structural property engineered from the ground up. Our results show that it is possible to defend perceptual models without adversarial training, without data augmentation. We believe this work is a first step toward making robustness an architectural default, rather than an afterthought — and we encourage the community to explore how design-first strategies can reshape the foundations of trustworthy image, video, and perceptual systems.

% \textbf{Future work.} This study primarily focuses on enhancing the robustness of convolutional IQA models. In future research, we plan to extend our approach to other architectures, including transformer-based models, to evaluate the generalizability of our defense method across different design paradigms.

% \textbf{Limitations.} First, the method requires retraining the model, which can be computationally demanding and may limit rapid deployment. Second, the integration of a 2D Fast Fourier Transform within the architecture introduces computational overhead, slowing down inference speed. Third, these modifications are specifically tailored for convolutional neural networks and are not directly applicable to transformer-based architectures, limiting the generalizability of the approach. Finally, the architecture necessitates the use of a pooling layer to reduce the input tensor dimensions to a square shape, imposing constraints on model design and flexibility. Recognizing these limitations provides important context for interpreting the results and suggests directions for future research to address these challenges.
\textbf{Limitations and Future Scope.} While our method demonstrates strong robustness gains, it comes with certain trade-offs. The incorporation of FFT-based transformations introduces moderate computational overhead, slightly increasing inference time. The method also requires light retraining after architectural changes, which may not suit all deployment scenarios. Additionally, our design is currently tailored for convolutional networks and has not yet been extended to transformer-based architectures. Finally, the need for square input tensors imposes minor constraints on architectural flexibility. However, we view these not as hard limitations, but as opportunities — to develop faster spectral approximations, to explore transformer-compatible architectural priors, and to build models where robustness is truly native by design.

% \begin{acks}
% \end{acks}

\bibliography{references}

\end{document}